\documentclass[pdflatex,sn-mathphys-num,iicol]{sn-jnl}

\usepackage{graphicx}%
\usepackage{multirow}%
\usepackage{amsmath,amssymb,amsfonts}%
\usepackage{amsthm}%
\usepackage{mathrsfs}%
\usepackage[title]{appendix}%
\usepackage{xcolor}%
\usepackage{textcomp}%
\usepackage{manyfoot}%
\usepackage{booktabs}%
\usepackage{xurl}%
\DeclareUrlCommand\burl{\urlstyle{sf}}%

\makeatletter
\long\def\@figurecaption#1#2{{\figurecaptionfont\leftskip\z@\rightskip\z@\parfillskip\z@ plus 1fil{\bfseries#1}\hskip.7em#2\par}}%
\long\def\@tablecaption#1#2{\setbox\tabcapbox\vbox{\tablecaptionfont\leftskip\z@\rightskip\z@\parfillskip\z@ plus 1fil{\bfseries #1}{\hskip2mm}#2\vphantom{y}\par}\box\tabcapbox}%
\makeatother

\begin{document}

\title[KeyWise: Efficient Video-Only Piano Transcription with a Lightweight Per-Key Architecture]{KeyWise: Efficient Video-Only Piano Transcription with a Lightweight Per-Key Architecture}

\author*[1]{\fnm{Anastasios} \sur{Chatzigiannis}}\email{achatzigiannis@uth.gr}
\author[1]{\fnm{Natalia} \sur{Panagou}}\email{npanagou@uth.gr}
\author[1]{\fnm{Maria} \sur{Koziri}}\email{mkoziri@uth.gr}

\affil*[1]{\orgdiv{Department of Informatics and Telecommunications}, \orgname{University of Thessaly}, \orgaddress{\city{Lamia}, \country{Greece}}}

\abstract{Audio-based piano transcription is hampered by the sustain pedal and by noisy or missing soundtracks, problems that video-only transcription avoids, since a top-view video shows directly when each key is pressed and released. Most visual transcription networks, however, describe the whole keyboard with one global feature vector and must learn where each key is. In this paper, we propose KeyWise, a lightweight per-key architecture that exploits the known geometry of the keyboard instead of learning it. The rectified keyboard is divided into 88 equal tiles that serve as regions of interest (ROIs), and one small recurrent network, shared by all keys, follows each key over time. On PianoVAM, KeyWise reaches the onset accuracy of the state-of-the-art V2N (95.0\% vs.\ 94.7\% onset F1 at 50~ms and 97.8\% vs.\ 97.0\% at 100~ms) with 1.3~M instead of 28.2~M parameters, and it runs 4.7 times faster on the same GPU. At 100~ms, it is also significantly more accurate when the key releases (97.0\% vs.\ 95.5\% F1) or the velocities (89.4\% vs.\ 84.5\% F1) must be correct as well. On PianoYT, KeyWise pre-trained on PianoVAM reaches 69.4\% onset F1 at 100~ms, comparable to the best published result, although it needs fine-tuning for new recording conditions.}


\maketitle

\section{Introduction}\label{sec:intro}

Automatic music transcription (AMT) is the process of (automatically) converting a music performance into a symbolic representation such as MIDI or a score~\cite{benetos2019}. Among all instruments, the piano is the most studied in AMT, both because of its central role in Western music and because large, precisely aligned datasets can be recorded on computer-controlled instruments such as the Yamaha Disklavier~\cite{hawthorne2018}. It is nevertheless a demanding case. Piano spans 88 pitches over more than seven octaves and is highly polyphonic, with dense chords and both hands often playing at once. Its notes are rich in harmonics that overlap between simultaneous pitches, which causes octave and harmonic confusions. Each note begins with a sharp percussive attack and then decays, so the start of a note is well defined but its end is acoustically ambiguous. This is mainly why deep audio models built on the onsets-and-frames formulation have brought onset detection close to saturation on clean studio recordings~\cite{hawthorne2018,kong2021}, while aspects such as offsets, velocities and robustness to unseen recording conditions remain open, and models trained on a single corpus generalize poorly to other pianos and rooms~\cite{martak2024}. A further difficulty is specific to the piano, namely that the sustain pedal decouples sound from action. When the pedal is held, a string keeps sounding long after its key has been released. As a result, audio cannot reliably recover how long each key was held, which is the information a performer's score and fingering depend on. Audio transcription also degrades in noisy or reverberant rooms, in multi-instrument mixes, and when the soundtrack is missing or corrupted~\cite{koepke2020,kim2026v2n}.

Video, on the other hand, observes the keys directly and is unaffected by all three aforementioned problems, namely the sustain pedal, the acoustic conditions and a missing soundtrack. For example, the moment a key is released shows whether a note that keeps sounding is sustained by the pedal. Which hand plays each note can guide towards a more accurate assignment of notes to staves. Moreover, video is useful when the audio is missing or has low quality, for example in silent videos, in noisy rooms, or in videos where other instruments play at the same time. However, video-only music transcription brings its own challenges, since hands occlude the keys, low frame rates limit temporal precision, and camera viewpoints and lighting vary. Despite the benefits visual information can bring to music transcription, and the widespread use of video in every aspect of life, including music performance, visual AMT remains underexplored, although it has recently begun to attract growing research interest.

Most of the recent visual transcription networks~\cite{koepke2020,su2020,zivanovic2025,kim2026v2n} describe the whole keyboard image with a single global feature vector and predict all 88 keys from it. Some of them add a position prior to their features~\cite{koepke2020,kim2026v2n}, but every key is still decided from one vector that describes the whole keyboard. Such a network has to learn two things at the same time, namely what a pressed key looks like and where each key is located. The second, however, is already given by the geometry of the keyboard itself. After a perspective warp, the keyboard becomes a straight strip, and the position of each key within the strip is known and fixed. In other words, pitch becomes, to a good approximation, position.

In this work, we propose KeyWise, a new model that is built around this observation. The rectified strip is divided into 88 equal tiles, one for each key, and each tile is used as the region of interest (ROI) of its key. The whole strip is processed by a convolutional encoder whose output has exactly one feature column per key ROI, so that the features of each key remain at the position of that key. A single small recurrent network, shared by all keys, then follows each key over time and decides, for every video frame, whether a note starts and whether the key is held. Since all 88 keys share the same weights, what the model learns from one key also benefits the others, including keys that are rarely played.

Our main contributions are summarized as follows:
\begin{itemize}
\item A new per-key model for video-only piano transcription, in which every key is a fixed ROI of the rectified keyboard and one shared network predicts the onset, the key state and the velocity of every key at every frame. The model has 1.3~M parameters, about 22 times fewer than the current state of the art, and runs 4.7 times faster on the same GPU.
\item Evidence that the per-key design matters. On the PianoVAM test set, the model reaches the onset accuracy of V2N~\cite{kim2026v2n} and is significantly more accurate at 100~ms when the key releases or the velocities are also considered. When the key ROIs are replaced by one global feature vector, the same network loses accuracy, mainly at the two ends of the keyboard, even when that vector keeps one position per key.
\end{itemize}

The rest of the paper is organized as follows. The related literature on automatic music transcription from video is reviewed in Section~\ref{sec:related}. In Section~\ref{sec:architecture} the proposed architecture is described. The experimental setup and the results are presented in Sections~\ref{sec:setup} and~\ref{sec:results}, respectively. Finally, Section~\ref{sec:conclusion} concludes the paper.

\section{Related work}\label{sec:related}

Before visual-only systems matured, video was mainly used to support an audio transcriber. In~\cite{wan2015} visual cues were used to resolve harmonic errors, while Lee et al.~\cite{lee2019} trained a two-stream audio-visual network, but only on simple material such as Hanon exercises. Wang et al.~\cite{wang2021} removed audio-predicted notes that would be physically impossible given the observed hand position, and the PianoVAM dataset paper~\cite{kim2025pianovam} uses video in the same filtering role. Li et al.~\cite{li2024} fused video and audio features through cross-attention; retrained alone on PianoVAM, its video branch detects onsets well but its notes have near-zero duration~\cite{kim2026v2n}. Nevertheless, all these systems still require a usable soundtrack and model offsets as the end of the sound rather than the key release.

The first visual-only systems relied on classical computer vision, detecting pressed keys from differences between the current frame and a background image of the empty keyboard~\cite{suteparuk2014}. Such methods are sensitive to illumination changes, shadows and occlusion by the hands. Other systems relied on depth cameras~\cite{rho2014,nisbet2017}, or classified cropped images of individual keys with a CNN after extensive pre-processing~\cite{akbari2018}.

More closely related to our work are the deep networks that transcribe the piano from video alone. Sight to Sound (S2S)~\cite{koepke2020} was the first end-to-end deep network for visual piano transcription. Given five consecutive grayscale frames of a registered $145 \times 800$ keyboard crop (25~fps, 0.2~s), it predicts which of the 88 notes have an onset near the middle frame. Its ResNet-18 backbone adds a $5 \times 1 \times 1$ 3D convolution that averages the early features of the five frames with learned weights, and a slope module that feeds a fixed vector of 88 linearly spaced values into the network to restore the horizontal position information that a classification CNN tends to discard. The same paper introduced the PianoYT dataset, which is also used in this work and described in Section~\ref{sec:data}. S2S reached 66.6\% onset F1 on the original PianoYT test split. However, S2S predicts onsets only, sees 0.2~s of context and supervises only the center frame. Video2Roll Net (V2R)~\cite{su2020}, part of a system that generates audio for silent piano videos, is a similar ResNet-18 that predicts the pressed keys from 5-frame windows.

Pay Attention to the Keys~\cite{zivanovic2025}, called PPAN in~\cite{kim2026v2n} and in this paper, replaces the CNN with a video Vision Transformer. It cuts six grayscale frames (0.2~s at 30~fps) of a $64 \times 784$ keyboard crop into $2 \times 16 \times 16$ spatio-temporal tubelets with absolute positional embeddings and processes them with a ViT-small encoder initialized from distilled video-ViT weights~\cite{wang2023}, on which the results depend strongly. PPAN also introduced R3, about 31 hours of rehearsal recordings by two professional pianists on two silent-system upright pianos, with synchronized video, audio and real MIDI. Trained on R3, PPAN outperforms S2S and V2R (onset F-score 0.87 vs.\ 0.84 on R3s) and is the only model with meaningful zero-shot transfer to PianoYT (0.40); after onset-only fine-tuning on PianoYT, it reaches 0.68, against 0.64 for both CNNs.

V2N~\cite{kim2026v2n} is the current state of the art and presents itself as the first complete visual transcription system, predicting onset, offset, key hold and velocity, with offsets defined as physical key releases. It clearly outperforms earlier systems on PianoVAM~\cite{kim2025pianovam}, as shown in Table~\ref{tab:vam}, it is the first visual system to report note-level velocity, and on R3 it outperforms all baselines by a wide margin (on R3s, 78.8\% vs.\ at most 44.0\% onset F1 at 50~ms). This comes at the cost of a large model, with 28.2~M parameters and 125~GFLOPs per 1~s clip. V2N turns each $800 \times 144$ grayscale frame (25~fps) into one vector with the S2S feature extractor, models 1~s windows with Conformer blocks, and predicts the four attributes with BiLSTM heads.

The aforementioned works differ from KeyWise in where the keys are represented. S2S and V2N add a position prior (a slope vector and a learned spatial encoding, respectively), but all the deep networks above predict the 88 keys jointly from features of the whole keyboard. Akbari et al.~\cite{akbari2018}, on the other hand, judged keys separately, but frame by frame and after hand-crafted pre-processing. KeyWise instead keeps one learned feature column per key from the encoder to the output and shares one temporal model over all keys, an idea related to the pitch-shared recurrent layers of the audio transcriber HPPNet~\cite{wei2022}. As a result, it is much smaller than the state of the art.

\section{KeyWise Architecture}\label{sec:architecture}

\begin{figure*}[t]
  \centering
  \includegraphics[width=0.95\textwidth]{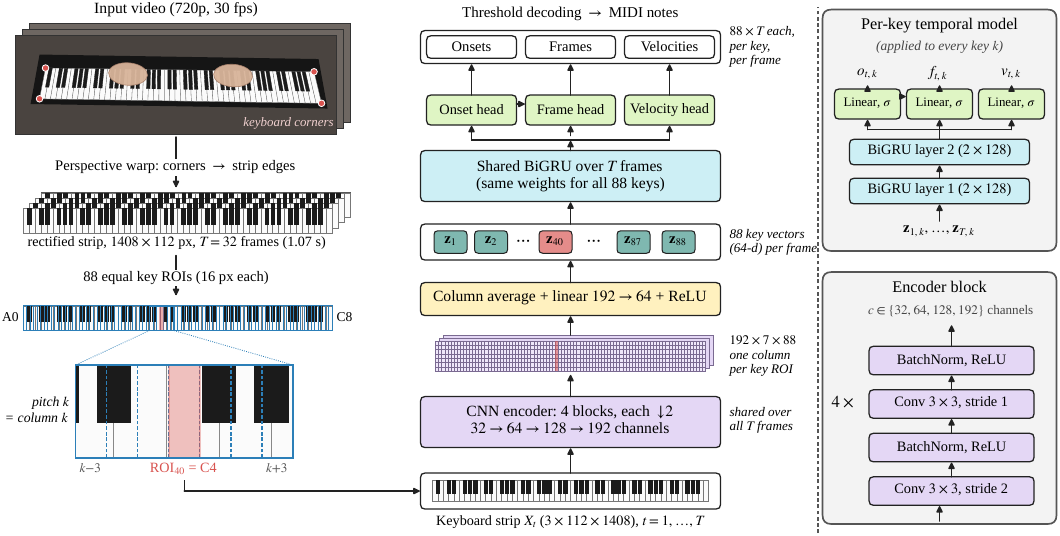}
  \caption{Overview of KeyWise. On the left, the keyboard is rectified from its four corners and cut into 88 equal key ROIs. In the middle, a CNN encoder gives one feature column per key ROI, and a BiGRU shared by all 88 keys feeds the onset, frame and velocity heads of every key at every frame. On the right, an encoder block and the per-key temporal model are shown in more detail.}
  \label{fig:architecture}
\end{figure*}

In most visual recognition problems, the objects of interest may appear anywhere in the image, at any scale and in any number. The 88 keys of a piano, in contrast, form a fixed and fully known layout, so once the keyboard is rectified, the position of each key, and therefore its pitch, is known in advance. KeyWise exploits this with a well-known tool of image and video processing, the region of interest (ROI), as in video coding, where important regions receive more bits~\cite{itti2004}. Here the ROIs are not detected but given by the instrument itself, as each key of the rectified keyboard is an ROI and the same decision process is applied to all 88 of them. Visual piano transcription thus becomes 88 identical per-key problems that share their parameters and differ only in where they look. Fig.~\ref{fig:architecture} gives an overview of the model.

\subsection{Keyboard rectification and key regions of interest}\label{sec:rect}

Each dataset gives the position of the keyboard, as four corner points per recording in PianoVAM and as a bounding box per video in PianoYT. The homography $\mathbf{M}$ that maps these four corners to the corners of a strip of width $W=1408$ and height $H=112$ pixels, with key A0 at the left and C8 at the right edge, gives the rectified color strip $X_t(u,v)=I_t\big(\pi(\mathbf{M}^{-1}[u,v,1]^{\top})\big)$, where $I_t$ is the video frame, sampled with bilinear interpolation, and $\pi([x,y,s]^{\top})=(x/s,y/s)$. PianoVAM videos are $1920\times1080$ at 60~fps, and PianoYT videos mostly $1920\times1080$ at 23.98--60~fps. We decode them at 720p, where the PianoVAM keyboard spans about 1150 pixels, and resample them to 30~fps by taking, for frame $t$, the native frame closest to time $t/30$~s.

\begin{figure}[t]
\centering
\includegraphics[width=\columnwidth]{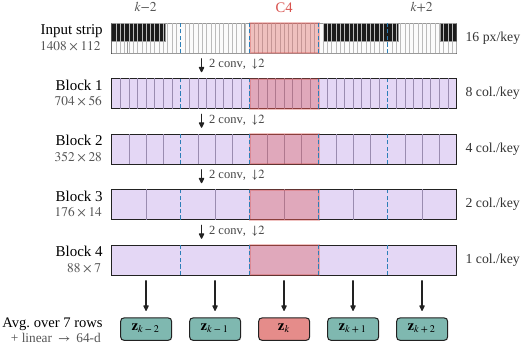}
\caption{Key ROIs through the encoder, for five keys around C4. Each block halves the width, so the 16-pixel ROI of key $k$ becomes exactly one column of the final feature map, which is averaged into the key vector $\mathbf{z}_k$ of Eq.~(\ref{eq:pool}).}
\label{fig:keyroi}
\end{figure}

We then cut the strip into 88 equal tiles, so that the ROI of key $k\in\{1,\dots,88\}$, counted from A0, is $R_k=\{(u,v): w(k-1)\le u<wk\}$, with $w=W/88=16$ pixels. The encoder reduces the width by a factor of 16, so it produces exactly one feature column per ROI, as illustrated in Fig.~\ref{fig:keyroi}. On a real keyboard, however, the black keys sit between equal white keys, so an equal tile does not follow the exact outline of its key. On a standard keyboard, the center of a tile lies up to about 17 pixels from the center of its key, and the center of the receptive field of encoder column $k$, which lies 7.5 pixels left of the tile center because of zero padding, up to about 20 pixels. Since the receptive field is 91 pixels wide, each key still lies well inside it, and the column can learn which pixels belong to its key. Indeed, as the analysis in Section~\ref{sec:res-vam} shows, the model almost never confuses neighboring keys.

\subsection{Encoder and shared temporal model}\label{sec:encoder}

The encoder $E$ has four blocks, each with a $3\times3$ convolution with stride 2 and a $3\times3$ convolution with stride 1, both followed by batch normalization~\cite{ioffe2015} and a ReLU. The blocks have 32, 64, 128 and 192 channels, and since each stride-2 convolution halves the height and the width, the $112\times1408$ strip becomes a feature map $F_t=E(X_t)\in\mathbb{R}^{192\times7\times88}$ of 7 rows and 88 columns, i.e.\ one column per key ROI. We average each column over its 7 rows and apply a shared $1\times1$ convolution with 64 output channels and a ReLU, which gives one vector per key and frame,
\begin{equation}
\mathbf{z}_{t,k}=\mathrm{ReLU}\Big(\mathbf{W}_p\,\frac{1}{7}\sum_{r=1}^{7}F_t[:,r,k]+\mathbf{b}_p\Big)\in\mathbb{R}^{64}.
\label{eq:pool}
\end{equation}
The same two-layer bidirectional GRU~\cite{cho2014}, with 128 units per direction and dropout 0.2, then runs over a clip of $T$ frames for each key separately, i.e.\ the 88 keys are 88 sequences with shared weights, and gives $\mathbf{h}_{t,k}\in\mathbb{R}^{256}$. Two linear heads give the onset and frame probabilities of key $k$ at frame $t$,
\begin{align}
o_{t,k}&=\sigma(\mathbf{w}_o^{\top}\mathbf{h}_{t,k}+b_o),\label{eq:heads}\\
f_{t,k}&=\sigma\big(\mathbf{w}_f^{\top}[\mathbf{h}_{t,k};\,\mathrm{sg}(o_{t,k})]+b_f\big),\nonumber
\end{align}
where $\sigma$ is the logistic function and $\mathrm{sg}(\cdot)$ stops the gradient. As in Onsets and Frames~\cite{hawthorne2018}, the frame head thus also receives the onset probability, without passing gradients to it. A third head, which predicts the velocity, is described in Section~\ref{sec:loss}. The model has 1,299,491 parameters, and almost all of its computation, about $1.87\times10^{9}$ multiply--accumulate operations (MACs) per frame or $5.6\times10^{10}$ per second of video at 30~fps, is spent in the encoder.

\subsection{Training objective and decoding}\label{sec:loss}

The targets are frame-level rolls. The onset target $y^{\mathrm{on}}_{t,k}$ is 1 at the frame nearest to a note onset of key $k$ and at the next frame, and the frame target $y^{\mathrm{fr}}_{t,k}$ is 1 from the onset to the offset of the note. Since positive targets are rare, we use a weighted binary cross-entropy
\begin{multline}
\ell(p,y;\lambda)=-\frac{1}{88\,T}\sum_{t,k}\big[\lambda\,y_{t,k}\log p_{t,k}\\
+(1-y_{t,k})\log(1-p_{t,k})\big],
\label{eq:bce}
\end{multline}
and the loss is $\mathcal{L}=\ell(o,y^{\mathrm{on}};8)+\ell(f,y^{\mathrm{fr}};2)$. If two onsets of the same key fall in consecutive frames, their onset targets merge, and only one note can be decoded. For velocity, a third linear head with a sigmoid output on $\mathbf{h}_{t,k}$ (257 parameters) predicts the MIDI velocity divided by 127. As in~\cite{hawthorne2018}, it is trained with the squared error at the onset frames only, added to $\mathcal{L}$ with weight 1, and a decoded note takes the velocity of its onset frame.

At test time, each video is processed in windows of 512 frames (about 17~s) that start every 448 frames, so consecutive windows overlap by 64 frames. In this overlap, we average the probabilities of the two windows, so that frames near the edge of a window, where the BiGRU sees little context, also benefit from the neighboring window. A note of key $k$ starts at frame $t$ when the onset probability crosses a threshold $\theta_{\mathrm{on}}$, i.e.\ $o_{t-1,k}<\theta_{\mathrm{on}}\le o_{t,k}$, and it ends at the first frame $t'>t$ with $f_{t',k}<\theta_{\mathrm{fr}}$ or with a new onset of the same key. The predicted onset is placed at the time stamp $t/30$~s of frame $t$, and onset frames also count as active, so every note lasts at least one frame. The two thresholds are chosen on validation data, as described in Section~\ref{sec:training}.

\section{Experimental setup}\label{sec:setup}

\subsection{Datasets}\label{sec:data}

\textbf{PianoVAM.} We use the official v1.1 split of PianoVAM~\cite{kim2025pianovam}, the same as V2N~\cite{kim2026v2n}, with 81 training recordings (we use 80, as recording 2024-09-05\_21-37-08 could not be decoded), 9 validation and 9 test recordings (42,241 test notes). As note offset we use the key release, which a camera can see, while the sound can last longer because of the sustain pedal. With our labels, the predictions released by V2N reproduce its published onset scores exactly and its other scores within 0.4 points.

\textbf{PianoYT.} PianoYT~\cite{koepke2020} consists of top-view YouTube piano videos, mostly by a single pianist, whose labels were obtained by running Onsets and Frames~\cite{hawthorne2018} on the audio of each video. These pseudo-labels carry the errors of the audio transcriber and pedal-extended offsets. The dataset lists 228 videos, 207 for training and 21 for testing. We removed seven unusable training videos (no camera view, a wrong keyboard box, or a duplicate) and kept 21 others for validation, which leaves 179 for training. We use all 21 test videos (33,535 notes), as in~\cite{zivanovic2025}; one of them is no longer on YouTube, and we evaluated it from an earlier copy. PPAN used all training videos still available (two were missing), so our training sets differ.

\subsection{Training and threshold selection}\label{sec:training}

KeyWise is implemented in PyTorch and trained on one NVIDIA RTX 3090 GPU, on clips of $T=32$ frames (1.07~s) in batches of 8. We use AdamW~\cite{loshchilov2019} with learning rate $6\times10^{-4}$ and weight decay $10^{-4}$, a linear warm-up over 3\% of the steps and a cosine decay~\cite{loshchilov2017} to 2\% of the peak learning rate, 25 epochs, gradient clipping at 3.0 and mixed precision. On 80\% of the clips, we apply random augmentation, namely horizontal scaling ($\pm0.6$\%), horizontal and vertical shifts ($\pm3$ pixels), rotation ($\pm0.3^\circ$), brightness and contrast changes ($\pm15$\%), conversion to grayscale (probability 0.1) and Gaussian noise (standard deviation 0.02, probability 0.3). We keep the epoch with the best frame-level onset F1 on the validation split. For fine-tuning on PianoYT, we start from a PianoVAM model and train for 12 epochs with learning rate $2\times10^{-4}$ and 5\% warm-up. Each setting (PianoVAM, PianoVAM with the velocity head, PianoYT, and PianoYT after PianoVAM pre-training) is trained three times with different random seeds, which change the initial weights, the order of the training clips and the augmentation.

The thresholds $\theta_{\mathrm{on}}$ and $\theta_{\mathrm{fr}}$ are chosen on the validation split in three steps. We first search $\theta_{\mathrm{on}}$ (from 0.05 to 0.98) with $\theta_{\mathrm{fr}}=0.5$, then $\theta_{\mathrm{fr}}$ (from 0.1 to 0.9) with the best $\theta_{\mathrm{on}}$, and finally $\theta_{\mathrm{on}}$ again with the best $\theta_{\mathrm{fr}}$. On PianoVAM we maximize the note F1 with onsets and offsets (with the default offset rule of mir\_eval~\cite{raffel2014}, for this choice only), and on PianoYT the onset F1, because its offsets are unreliable, in both cases at 50~ms onset tolerance. The test split is never used for any choice. The selected onset threshold is 0.90 for all three PianoVAM runs and 0.50--0.60 on PianoYT, and the frame threshold is 0.50--0.70 on PianoVAM and 0.60--0.80 on PianoYT. With a positive weight $\lambda$ in Eq.~(\ref{eq:bce}), the model outputs $\lambda/(1+\lambda)$ where an unweighted model would output 0.5, i.e.\ 0.89 for onsets and 0.67 for frames, so $\theta_{\mathrm{on}}=0.90$ follows from the loss weight.

\subsection{Evaluation}\label{sec:eval}

We evaluate all systems with the note metrics of mir\_eval~\cite{raffel2014}, which are used by all the works reviewed in Section~\ref{sec:related}; our own scorer, which matches the notes of each pitch separately for speed, gives exactly the same results. A predicted note counts as correct for the onset F1 if its pitch is correct and its onset lies within $\pm50$~ms or $\pm100$~ms of a reference note, and each reference note can be matched only once. The 50~ms tolerance is the standard in audio transcription, but at video frame rates a frame already lasts 33--40~ms, which is why visual works also report results at 100~ms. Following V2N~\cite{kim2026v2n}, we also report the onset-and-offset F1 (``+Off''), in which the offset must also lie within the tolerance, and the frame F1 on a 60~Hz grid. The velocity metrics ``+Vel'' and ``+Off+Vel'' additionally require a correct velocity, which, following the code of V2N, means a velocity within 12.7 MIDI steps of the reference. This rule is lenient, since even a constant velocity scores 62.7--64.2\% +Vel at 50~ms. All scores are averaged over the recordings.

For KeyWise, we report the mean and standard deviation of three runs. To compare two systems, we average the score of each recording over the runs and apply the exact two-sided Wilcoxon signed-rank test~\cite{wilcoxon1945} to the paired recordings; with only 9 test recordings, the smallest possible $p$-value is 0.004. As several metrics are compared, we also correct the $p$-values with the Holm--Bonferroni method~\cite{holm1979}, separately for the five note and the four velocity metrics, and we compute 95\% bootstrap confidence intervals over the recordings. The MIDI files released by V2N for the PianoVAM test set are scored with exactly the same code as our own predictions, although the two systems receive different inputs (a $1408\times112$ color strip at 30~fps vs.\ an $800\times144$ grayscale image at 25~fps), which we regard as part of each design. For all other systems, we copy the numbers reported in~\cite{kim2026v2n} for PianoVAM and in~\cite{zivanovic2025} for PianoYT, on the same 21 test videos.

\subsection{Ablation variants}\label{sec:global}

To measure the effect of the key ROIs, we also train a global variant, which follows the conventional design. It uses the same encoder, but its feature map is pooled to 16 horizontal bins, flattened (3072 values) and projected to one 256-dimensional vector per frame, which is followed by a two-layer bidirectional GRU with 256 units per direction and two linear heads with 88 outputs. The global variant has 3,698,128 parameters and about the same computation as KeyWise ($1.83\times10^{9}$ MACs per frame). To separate the effect of the horizontal resolution from that of the shared weights, we also build Global-88, which pools the feature map to 88 bins, i.e.\ one feature column per key as in KeyWise, but then flattens them (16,896 values) and continues like the global variant, so that every key keeps its own weights (7,237,072 parameters, about the same computation). Both variants are trained three times, with the same data, training and decoding as KeyWise.

\section{Results}\label{sec:results}

In this section, we evaluate KeyWise from several aspects. First, in Section~\ref{sec:res-vam}, KeyWise is compared against the state of the art on PianoVAM and on PianoYT, and we also take a closer look at how accurately it locates the keys and times their onsets and releases. Then, in Section~\ref{sec:res-abl}, an ablation study replaces the key ROIs by a global feature vector, in order to assess the contribution of the per-key design. Finally, Section~\ref{sec:res-sum} discusses our main findings.

\subsection{Comparison with the state of the art}\label{sec:res-vam}

\begin{figure*}[t]
\centering
\includegraphics[width=\textwidth]{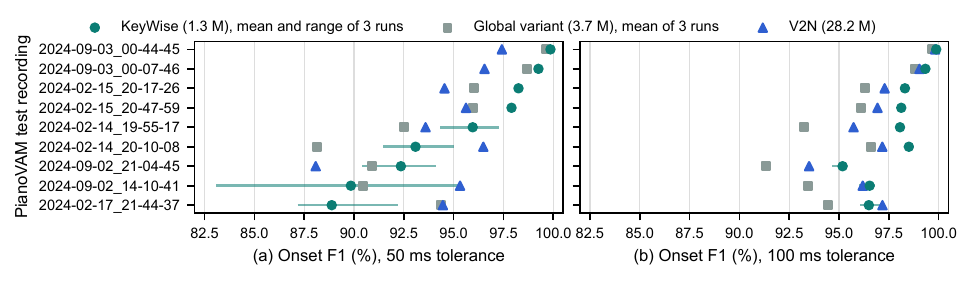}
\caption{Onset F1 per PianoVAM test recording at 50~ms (a) and 100~ms (b). For KeyWise, the mean and the range of three runs are shown, for the global variant the mean of three runs, and for V2N its released predictions. The x-axis starts at 82\%.}\label{fig:recs}
\end{figure*}

Table~\ref{tab:vam} summarizes the results on the PianoVAM test set. KeyWise reaches the onset accuracy of V2N with 22 times fewer parameters. At 50~ms, the two systems score 95.0\% and 94.7\% onset F1, and the difference is not significant ($p=1.0$). At 100~ms, KeyWise scores 97.8\% against 97.0\% and is better on 8 of the 9 recordings (Fig.~\ref{fig:recs}; mean difference 0.8 points, 95\% confidence interval 0.3--1.4), but after the Holm--Bonferroni correction this difference is not significant either ($p_{\mathrm{Holm}}=0.11$). The advantage of KeyWise becomes clear when the key release must also be found. With the offset inside the tolerance, KeyWise is better on all 9 recordings at 100~ms (97.0\% vs.\ 95.5\%, $p_{\mathrm{Holm}}=0.020$), while at 50~ms and in frame F1 the differences are not significant.

Adding the velocity head changes the note scores of KeyWise by less than 0.6 points. This head, with only 257 parameters, estimates velocities more accurately than V2N. At 100~ms, KeyWise is better on all 9 recordings in both +Vel (89.4\% vs.\ 84.5\%) and +Off+Vel (88.6\% vs.\ 83.2\%), with $p_{\mathrm{Holm}}=0.016$ in both cases, while at 50~ms it is ahead but not significantly. Its predicted velocities also correlate better with the reference ones ($r=0.83$--$0.85$ in the three runs vs.\ 0.77 for V2N). Finally, the video branch of Li et al.\ reaches 97.4\% onset F1 at 100~ms~\cite{kim2026v2n}, only 0.4 points below KeyWise, but its notes have almost no duration.

\begin{table*}[t]
\caption{F1 (\%) on the PianoVAM test set (9 recordings, 42,241 notes) at 50 and 100~ms onset tolerance. The last five rows are scored with the same code, and for KeyWise the mean $\pm$ standard deviation over three runs is given.}\label{tab:vam}%
\footnotesize\setlength{\tabcolsep}{1.5pt}
\begin{tabular*}{\textwidth}{@{\extracolsep\fill}lcccccccccc@{}}
\toprule
 & & & \multicolumn{2}{c}{Onset} & \multicolumn{2}{c}{+Off} & \multicolumn{2}{c}{+Vel} & \multicolumn{2}{c}{+Off+Vel} \\
\cmidrule{4-5}\cmidrule{6-7}\cmidrule{8-9}\cmidrule{10-11}
Model & Params & Frame & 50 & 100 & 50 & 100 & 50 & 100 & 50 & 100 \\
\midrule
S2S~\cite{koepke2020}$^{a}$ & -- & 41.7 & 60.1 & 94.5 & 21.8 & 49.6 & -- & -- & -- & -- \\
V2R~\cite{su2020}$^{a}$ & -- & 82.2 & 86.5 & 89.8 & 55.4 & 81.8 & -- & -- & -- & -- \\
Li et al.~\cite{li2024}$^{a,b}$ & --& (25.2) & 94.2 & 97.4 & (21.0) & (40.0) & -- & -- & -- & -- \\
PPAN~\cite{zivanovic2025}$^{a}$ & -- & 80.4 & 84.9 & 94.1 & 45.9 & 82.8 & -- & -- & -- & -- \\
V2N~\cite{kim2026v2n}, published & 28.2 M & 90.9 & 94.7 & 97.0 & 89.5 & 95.5 & 82.8 & 84.5 & 78.3 & 83.2 \\
\midrule
V2N, released MIDI & 28.2 M & 90.5 & 94.7 & 97.0 & 89.6 & 95.5 & 82.8 & 84.5 & 78.4 & 83.2 \\
Global variant & 3.7 M & $90.0_{\pm0.2}$ & $94.1_{\pm0.5}$ & $95.6_{\pm0.4}$ & $86.9_{\pm0.3}$ & $93.9_{\pm0.2}$ & -- & -- & -- & -- \\
Global-88 & 7.2 M & $89.1_{\pm0.5}$ & $92.4_{\pm1.5}$ & $95.2_{\pm0.4}$ & $83.9_{\pm2.1}$ & $93.3_{\pm0.5}$ & -- & -- & -- & -- \\
KeyWise & 1.3 M & $91.4_{\pm0.3}$ & $95.0_{\pm0.9}$ & $97.8_{\pm0.2}$ & $89.9_{\pm1.1}$ & $\mathbf{97.0}_{\pm0.2}$ & -- & -- & -- & -- \\
KeyWise + velocity & 1.3 M & $91.5_{\pm0.1}$ & $95.0_{\pm0.2}$ & $98.0_{\pm0.1}$ & $89.3_{\pm0.3}$ & $\mathbf{97.2}_{\pm0.1}$ & $86.8_{\pm0.6}$ & $\mathbf{89.4}_{\pm0.5}$ & $81.7_{\pm0.5}$ & $\mathbf{88.6}_{\pm0.6}$ \\
\botrule
\end{tabular*}
\par\vspace{3pt}{\footnotesize $^{a}$Retrained and reported by~\cite{kim2026v2n}; values in parentheses come from heads that~\cite{kim2026v2n} marks as degenerate. $^{b}$Video branch. +Vel requires a correct onset and velocity, and +Off+Vel also a correct offset, as defined in Section~\ref{sec:eval}. Bold values are significantly better than V2N (released MIDI) after the Holm--Bonferroni correction.\par}
\end{table*}

Unlike V2N, which uses a fixed threshold of 0.5, we choose the thresholds on the validation split, where a threshold of 0.5 would lower the onset F1 of KeyWise at 50~ms by 2.4--5.3 points. On the test set, however, the effect on onsets is small. For two of the three runs, a threshold of 0.5 gives 94.1\% and 97.7\% onset F1 at 50 and 100~ms, instead of 95.0\% and 97.8\%, while the +Off F1 at 50~ms drops from 89.6\% to 84.6\%. The thresholds $\lambda/(1+\lambda)$ give practically the same scores, and keeping the last instead of the best epoch lowers only the onset F1 at 50~ms, by 1.4 points.

\textbf{Key localization and timing.} Here we analyze the first run, the median of the three. Wrong keys are very rare, since only 15 missed and 5 extra notes are confused with a key one or two semitones or an octave away, so the rectification and the key ROIs locate the keys correctly. The recall of a key falls slightly as its tile center moves away from its real center, but so does the recall of V2N, which has no tiles, so this effect follows the layout of the keys rather than the tiles. More than half of the errors at 50~ms are timing errors. Black keys are detected later (median onset error $+15.6$ vs.\ $+3.9$~ms) and slightly less often than white keys (recall 92.8\% vs.\ 97.3\%), probably because their movement is smaller and harder to see from above. The timing error is almost constant within a recording but varies between recordings, from $-5.5$ to $+30.5$~ms, while V2N places its onsets early, by a median of 7--12~ms. Compared with V2N, KeyWise misses fewer notes with no prediction nearby (916 vs.\ 1,333) and adds fewer spurious ones (268 vs.\ 644), with a similar number of timing errors (826 vs.\ 910 missed notes). It also finds the key releases well, since counting only the offsets gives an F1 similar to that of V2N at 50~ms (91.5--92.8\% in the three runs vs.\ 92.1\%) and higher at 100~ms (97.0--97.4\% vs.\ 95.7\%). Finally, KeyWise finds 72.1\% of the notes that repeat the same key within 150~ms, against 62.1\% for V2N.

\textbf{PianoYT.} The results on PianoYT are given in Table~\ref{tab:yt}. Trained on PianoYT only, KeyWise reaches 66.7\% onset F1 at 100~ms, whereas KeyWise trained only on PianoVAM and applied to PianoYT without further training reaches only 15.3\%, even with thresholds chosen on the PianoYT validation split. V2N reports a similar collapse between PianoVAM and R3~\cite{kim2026v2n}, while the zero-shot result of PPAN (0.40) starts from R3 and from large-scale video pre-training. When KeyWise is pre-trained on PianoVAM and then fine-tuned on PianoYT, it reaches 69.4\%. Pre-training thus raises the onset F1 by 2.8 and 2.6 points at 50 and 100~ms, and every fine-tuned run is above every run trained on PianoYT only, although the per-video difference is not significant ($p=0.065$ and $p=0.060$). The fine-tuned runs (68.4\% to 69.9\%) are comparable to the best published result, 68\% by PPAN~\cite{zivanovic2025}, which is reported with two digits, although this comparison is not paired and the training data differ. All scores at 50~ms are much lower, partly because the labels of PianoYT come from an audio transcriber and their timing is less reliable.

\begin{table}[t]
\caption{Onset F1 (\%) on the PianoYT test set (21 videos, 33,535 notes). For KeyWise trained on PianoYT, the mean $\pm$ standard deviation over three runs is given.}\label{tab:yt}%
\setlength{\tabcolsep}{3pt}
\begin{tabular*}{\columnwidth}{@{\extracolsep\fill}llcc@{}}
\toprule
Model & Extra data & 50~ms & 100~ms \\
\midrule
S2S \cite{koepke2020}$^{a}$ & R3 & -- & 64 \\
V2R \cite{su2020}$^{a}$ & R3 & -- & 64 \\
PPAN \cite{zivanovic2025} & video + R3 & -- & 68 \\
\midrule
KeyWise & none & $41.9{\pm}1.6$ & $66.7{\pm}1.4$ \\
KeyWise, zero-shot$^{b}$ & PianoVAM & 6.5 & 15.3 \\
KeyWise, fine-tuned & PianoVAM & $44.7{\pm}0.9$ & $69.4{\pm}0.9$ \\
\botrule
\end{tabular*}
\footnotetext[a]{Retrained and reported by \cite{zivanovic2025} on the same 21 test videos. $^{b}$No training on PianoYT; one run. ``Video'' refers to VideoMAE~V2 pre-training~\cite{wang2023}.}
\end{table}

\subsection{Ablation study: key ROIs versus a global feature vector}\label{sec:res-abl}

Table~\ref{tab:abl} compares KeyWise with the two ablation variants described in Section~\ref{sec:global}. We first consider the global variant, which uses the same encoder, data, training and decoding as KeyWise, but merges the features into one global vector and has 2.8 times more parameters. At 50~ms, the difference between the two is not significant, but at 100~ms KeyWise is better on all 9 recordings (97.8\% vs.\ 95.6\%, $p_{\mathrm{Holm}}=0.020$) and in each of the three pairs of runs. It is also worth noting that the global variant falls below V2N at 100~ms on all 9 recordings, which shows that keeping the features per key ROI is what lifts the same encoder above V2N.

\begin{table}[t]
\caption{KeyWise, the global variant and Global-88 (G-88) on PianoVAM. The upper part gives the mean F1 (\%) of three runs and the Holm--Bonferroni corrected Wilcoxon $p$-values against the global variant ($p_1$) and Global-88 ($p_2$), and the lower part the notes found and the errors at 50~ms of the first run, out of the 42,241 reference notes of the test set.}\label{tab:abl}%
\footnotesize\setlength{\tabcolsep}{2.2pt}
\begin{tabular}{@{}lrrrrr@{}}
\toprule
 & KeyWise & Global & G-88 & $p_1$ & $p_2$ \\
\midrule
Parameters & 1.3 M & 3.7 M & 7.2 M & & \\
Onset F1, 50 ms & 95.0 & 94.1 & 92.4 & 0.250 & 0.078 \\
Onset F1, 100 ms & 97.8 & 95.6 & 95.2 & 0.020 & 0.020 \\
+Off F1, 50 ms & 89.9 & 86.9 & 83.9 & 0.148 & 0.078 \\
+Off F1, 100 ms & 97.0 & 93.9 & 93.3 & 0.020 & 0.020 \\
Frame F1 & 91.4 & 90.0 & 89.1 & 0.117 & 0.059 \\
\midrule
Notes found & 40,484 & 39,568 & 39,203 & & \\
Missed, timing & 826 & 877 & 1,284 & & \\
Missed, wrong key & 15 & 43 & 44 & & \\
Missed, no note nearby & 916 & 1,753 & 1,710 & & \\
Extra, timing & 824 & 887 & 1,285 & & \\
Extra, wrong key & 5 & 23 & 30 & & \\
Extra, no note nearby & 268 & 980 & 1,078 & & \\
\botrule
\end{tabular}
\end{table}

\begin{figure}[t]
\centering
\includegraphics[width=\columnwidth]{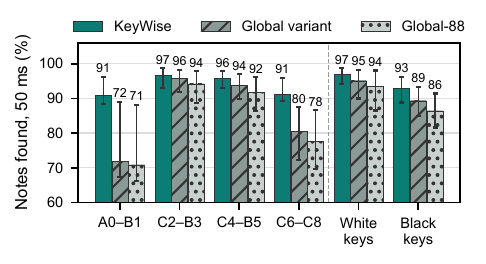}
\caption{Reference notes found at 50~ms on the PianoVAM test set, by register and key color, with the 9 recordings pooled and the mean of three runs shown. Error bars show 95\% bootstrap confidence intervals over the recordings.}\label{fig:reg}
\end{figure}

The errors at 50~ms illustrate where this difference comes from. As the lower part of Table~\ref{tab:abl} shows, both models make about the same number of timing errors, but the global variant misses about twice as many notes with no prediction nearby and predicts 3.7 times more notes where no note was played. As Fig.~\ref{fig:reg} shows, these losses are concentrated at the two ends of the keyboard. In the lowest register (A0--B1, 15 keys), the global variant finds 72\% of the notes, against 91\% for KeyWise (difference 19 points, 95\% confidence interval 9--23), and in the highest octaves (C6--C8) 80\% against 91\%. This is to be expected, since in the global variant every key has its own output weights, and the keys at the two ends are played rarely (the lowest register has about 1\% of the test notes), so their weights get little training. In KeyWise, on the other hand, all key ROIs share the same weights, so the lowest key also learns from the notes played in the middle of the keyboard. Indeed, the gain of KeyWise over the global variant grows as the number of training notes of a key falls (Spearman $\rho=-0.75$ over the 73 keys with test notes).

Global-88 tests whether this gain comes from the 88 positions or from the shared weights. Although it keeps one feature column per key, it is slightly worse than the global variant on all five metrics, and KeyWise is better than Global-88 at 100~ms on all 9 recordings ($p_{\mathrm{Holm}}=0.020$). Global-88 also loses most at the two ends of the keyboard, where it finds 71\% and 78\% of the notes against 91\% for KeyWise, and its gap to KeyWise also grows as the number of training notes of a key falls ($\rho=-0.79$). Therefore, the finer resolution alone does not explain the gain of KeyWise, which points to the weights shared by all key ROIs as its source.

\subsection{Discussion}\label{sec:res-sum}

Summarizing our findings, using the geometry of the keyboard instead of learning it clearly pays off. The network only needs to find a key within a known neighborhood, and sharing one temporal model between all key ROIs helps the rarely played keys while keeping the model small.

As far as computation is concerned, the gain of KeyWise is smaller than its reduction in parameters, because almost all of its computation is spent in the encoder. At test time, V2N processes every frame twice in overlapping 1~s clips and KeyWise about 1.14 times in overlapping windows, so KeyWise needs 3.9 times fewer operations per second of video than V2N, and 2.2 times fewer without overlap in either system. On the same RTX 3090 GPU, in 32-bit precision, it needs 15~ms per second of video against 70~ms for V2N (13 and 35~ms without overlap), V2N is 3.6 and 5.3 times slower in bfloat16 and float16, and with video decoding and rectification the whole KeyWise pipeline runs 2.3 times faster than real time. Finally, KeyWise currently relies on the keyboard corners given by the datasets and on a fully visible 88-key keyboard, it needs fine-tuning for new recording conditions like earlier visual systems, and its keys exchange no information after the encoder, which leaves room for models that also take neighboring keys into account.

\section{Conclusion}\label{sec:conclusion}

We proposed KeyWise, a video-only piano transcription model that exploits the fixed geometry of the keyboard, in which every key is a fixed ROI of the rectified keyboard and one small network, shared by all keys, predicts their onsets, states and velocities. With only 1.3~M parameters, KeyWise reaches the onset accuracy of the state-of-the-art V2N, is significantly more accurate at 100~ms when key releases or velocities are also considered, and runs 4.7 times faster. The ablation study clearly indicated that the per-key design is what makes the difference, since a global feature vector made the model worse, mostly at the rarely played ends of the keyboard, even when it kept one position per key. On PianoYT, whose YouTube videos vary in recording conditions and carry audio-derived labels, pre-training on PianoVAM led to 69.4\% onset F1 at 100~ms, comparable to the best published result. In future work, we plan to take advantage of more of the visual information in a performance video, such as the hands of the pianist, on our way towards a symbolic score.

\backmatter

\section*{Statements and Declarations}

\begin{itemize}
\item \textbf{Funding.} The authors did not receive support from any organization for the submitted work.
\item \textbf{Competing interests.} The authors declare no competing interests.
\item \textbf{Ethics approval and consent to participate.} Not applicable.
\item \textbf{Consent for publication.} Not applicable.
\item \textbf{Data availability.} The PianoVAM~\cite{kim2025pianovam} and PianoYT~\cite{koepke2020} datasets are publicly available from their authors. One PianoYT test video, no longer on YouTube, was evaluated from an earlier copy. The lists of removed and validation videos, the predicted MIDI files and the scores per recording are available from the corresponding author on reasonable request.
\item \textbf{Materials availability.} Not applicable.
\item \textbf{Code availability.} The code is available from the corresponding author on reasonable request.
\item \textbf{Use of AI tools.} An AI assistant (Claude, Anthropic) was used to assist with code development and edit parts of the manuscript. The authors designed the method and experiments, verified all results and take full responsibility for the content.
\item \textbf{Author contribution.} A.C.: conceptualization, methodology, software, experiments, formal analysis, writing of the original draft. N.P.: supervision, methodology, writing (review and editing). M.K.: supervision, writing (review and editing). All authors read and approved the final manuscript.
\end{itemize}

\bibliography{references}

@article{benetos2019,
  author    = "Benetos, E. and Dixon, S. and Duan, Z. and Ewert, S.",
  title     = "Automatic music transcription: an overview",
  journal   = "IEEE Signal Processing Magazine",
  volume    = "36",
  number    = "1",
  pages     = "20--30",
  year      = "2019",
  doi       = "10.1109/MSP.2018.2869928"
}

@inproceedings{hawthorne2018,
  author    = "Hawthorne, C. and Elsen, E. and Song, J. and Roberts, A. and Simon, I. and Raffel, C. and Engel, J. and Oore, S. and Eck, D.",
  title     = "Onsets and frames: dual-objective piano transcription",
  booktitle = "Proceedings of the 19th International Society for Music Information Retrieval Conference (ISMIR)",
  pages     = "50--57",
  year      = "2018",
  doi       = "10.5281/zenodo.1492341"
}

@article{kong2021,
  author    = "Kong, Q. and Li, B. and Song, X. and Wan, Y. and Wang, Y.",
  title     = "High-resolution piano transcription with pedals by regressing onset and offset times",
  journal   = "IEEE/ACM Transactions on Audio, Speech, and Language Processing",
  volume    = "29",
  pages     = "3707--3717",
  year      = "2021",
  doi       = "10.1109/TASLP.2021.3121991"
}

@inproceedings{martak2024,
  author    = "Mart{\'a}k, L. S. and Hu, P. and Widmer, G.",
  title     = "Quantifying the corpus bias problem in automatic music transcription systems",
  booktitle = "Proceedings of the 1st International Workshop on Sound Signal Processing Applications (IWSSPA)",
  year      = "2024",
  doi       = "10.48550/arXiv.2408.04737"
}

@article{akbari2018,
  author    = "Akbari, M. and Liang, J. and Cheng, H.",
  title     = "A real-time system for online learning-based visual transcription of piano music",
  journal   = "Multimedia Tools and Applications",
  volume    = "77",
  number    = "19",
  pages     = "25513--25535",
  year      = "2018",
  doi       = "10.1007/s11042-018-5803-1"
}

@techreport{suteparuk2014,
  author      = "Suteparuk, P.",
  title       = "Detection of piano keys pressed in video",
  institution = "Department of Computer Science, Stanford University",
  year        = "2014",
  url         = "https://stacks.stanford.edu/file/druid:bf950qp8995/Suteparuk.pdf"
}

@inproceedings{rho2014,
  author    = "Rho, S. and Hwang, J.-I. and Kim, J.",
  title     = "Automatic piano tutoring system using consumer-level depth camera",
  booktitle = "Proceedings of the IEEE International Conference on Consumer Electronics (ICCE)",
  pages     = "3--4",
  year      = "2014",
  doi       = "10.1109/ICCE.2014.6775884"
}

@misc{nisbet2017,
  author       = "Nisbet, A. and Green, R.",
  title        = "Capture of dynamic piano performance with depth vision",
  howpublished = "Department of Computer Science, University of Canterbury",
  year         = "2017",
  url          = "https://albertnis.com/resources/2017-05-10-piano-vision/Nisbet_Green_Capture_of_Dynamic_Piano%20Performance_with_Depth_Vision.pdf"
}

@article{wan2015,
  author    = "Wan, Y. and Wang, X. and Zhou, R. and Yan, Y.",
  title     = "Automatic piano music transcription using audio-visual features",
  journal   = "Chinese Journal of Electronics",
  volume    = "24",
  number    = "3",
  pages     = "596--603",
  year      = "2015",
  doi       = "10.1049/cje.2015.07.027"
}

@inproceedings{lee2019,
  author    = "Lee, J. and Doosti, B. and Gu, Y. and Cartledge, D. and Crandall, D. J. and Raphael, C.",
  title     = "Observing pianist accuracy and form with computer vision",
  booktitle = "Proceedings of the IEEE Winter Conference on Applications of Computer Vision (WACV)",
  pages     = "1505--1513",
  year      = "2019",
  doi       = "10.1109/WACV.2019.00165"
}

@inproceedings{wang2021,
  author    = "Wang, X. and Xu, W. and Liu, J. and Yang, W. and Cheng, W.",
  title     = "An audio-visual fusion piano transcription approach based on strategy",
  booktitle = "Proceedings of the 24th International Conference on Digital Audio Effects (DAFx20in21)",
  pages     = "308--315",
  year      = "2021",
  url       = "https://dafx2020.mdw.ac.at/Proceedings/papers/DAFx20in21_paper_19.pdf"
}

@inproceedings{koepke2020,
  author    = "Koepke, A. S. and Wiles, O. and Moses, Y. and Zisserman, A.",
  title     = "Sight to sound: an end-to-end approach for visual piano transcription",
  booktitle = "Proceedings of the IEEE International Conference on Acoustics, Speech and Signal Processing (ICASSP)",
  pages     = "1838--1842",
  year      = "2020",
  doi       = "10.1109/ICASSP40776.2020.9053115"
}

@inproceedings{su2020,
  author    = "Su, K. and Liu, X. and Shlizerman, E.",
  title     = "Audeo: audio generation for a silent performance video",
  booktitle = "Advances in Neural Information Processing Systems 33 (NeurIPS)",
  year      = "2020",
  doi       = "10.48550/arXiv.2006.14348"
}

@inproceedings{zivanovic2025,
  author    = "Zivanovic, U. and Pilkov, I. and Cancino-Chac{\'o}n, C.",
  title     = "Pay attention to the keys: visual piano transcription using transformers",
  booktitle = "Proceedings of the Thirty-Fourth International Joint Conference on Artificial Intelligence (IJCAI)",
  pages     = "10243--10251",
  year      = "2025",
  doi       = "10.24963/ijcai.2025/1138"
}

@inproceedings{kim2025pianovam,
  author    = "Kim, Y. and Park, J. and Bae, J. and Kim, K. and Kwon, T. and Lerch, A. and Nam, J.",
  title     = "{PianoVAM}: a multimodal piano performance dataset",
  booktitle = "Proceedings of the 26th International Society for Music Information Retrieval Conference (ISMIR)",
  year      = "2025",
  doi       = "10.48550/arXiv.2509.08800"
}

@inproceedings{kim2026v2n,
  author    = "Kim, Y. and Sohn, H. and Nam, J. and Lerch, A.",
  title     = "Multi-task multi-frame visual piano transcription",
  booktitle = "Proceedings of the 27th International Society for Music Information Retrieval Conference (ISMIR)",
  year      = "2026, to appear",
  doi       = "10.48550/arXiv.2608.03419"
}

@article{li2024,
  author    = "Li, Y. and Wang, X. and Wu, R. and Xu, W. and Cheng, W.",
  title     = "A two-stage audio-visual fusion piano transcription model based on the attention mechanism",
  journal   = "IEEE/ACM Transactions on Audio, Speech, and Language Processing",
  volume    = "32",
  pages     = "3618--3630",
  year      = "2024",
  doi       = "10.1109/TASLP.2024.3426303"
}

@inproceedings{wang2023,
  author    = "Wang, L. and Huang, B. and Zhao, Z. and Tong, Z. and He, Y. and Wang, Y. and Wang, Y. and Qiao, Y.",
  title     = "{VideoMAE V2}: scaling video masked autoencoders with dual masking",
  booktitle = "Proceedings of the IEEE/CVF Conference on Computer Vision and Pattern Recognition (CVPR)",
  pages     = "14549--14560",
  year      = "2023",
  doi       = "10.1109/CVPR52729.2023.01398"
}

@inproceedings{wei2022,
  author    = "Wei, W. and Li, P. and Yu, Y. and Li, W.",
  title     = "{HPPNet}: modeling the harmonic structure and pitch invariance in piano transcription",
  booktitle = "Proceedings of the 23rd International Society for Music Information Retrieval Conference (ISMIR)",
  pages     = "709--716",
  year      = "2022",
  doi       = "10.5281/zenodo.7316762"
}

@article{itti2004,
  author    = "Itti, L.",
  title     = "Automatic foveation for video compression using a neurobiological model of visual attention",
  journal   = "IEEE Transactions on Image Processing",
  volume    = "13",
  number    = "10",
  pages     = "1304--1318",
  year      = "2004",
  doi       = "10.1109/TIP.2004.834657"
}

@inproceedings{ioffe2015,
  author    = "Ioffe, S. and Szegedy, C.",
  title     = "Batch normalization: accelerating deep network training by reducing internal covariate shift",
  booktitle = "Proceedings of the 32nd International Conference on Machine Learning (ICML)",
  pages     = "448--456",
  year      = "2015",
  doi       = "10.48550/arXiv.1502.03167"
}

@inproceedings{cho2014,
  author    = "Cho, K. and {van Merri{\"e}nboer}, B. and Gulcehre, C. and Bahdanau, D. and Bougares, F. and Schwenk, H. and Bengio, Y.",
  title     = "Learning phrase representations using {RNN} encoder--decoder for statistical machine translation",
  booktitle = "Proceedings of the 2014 Conference on Empirical Methods in Natural Language Processing (EMNLP)",
  pages     = "1724--1734",
  year      = "2014",
  doi       = "10.3115/v1/D14-1179"
}

@inproceedings{loshchilov2019,
  author    = "Loshchilov, I. and Hutter, F.",
  title     = "Decoupled weight decay regularization",
  booktitle = "International Conference on Learning Representations (ICLR)",
  year      = "2019",
  doi       = "10.48550/arXiv.1711.05101"
}

@inproceedings{loshchilov2017,
  author    = "Loshchilov, I. and Hutter, F.",
  title     = "{SGDR}: stochastic gradient descent with warm restarts",
  booktitle = "International Conference on Learning Representations (ICLR)",
  year      = "2017",
  doi       = "10.48550/arXiv.1608.03983"
}

@inproceedings{raffel2014,
  author    = "Raffel, C. and McFee, B. and Humphrey, E. J. and Salamon, J. and Nieto, O. and Liang, D. and Ellis, D. P. W.",
  title     = "mir\_eval: a transparent implementation of common {MIR} metrics",
  booktitle = "Proceedings of the 15th International Society for Music Information Retrieval Conference (ISMIR)",
  pages     = "367--372",
  year      = "2014",
  doi       = "10.5281/zenodo.1416528"
}

@article{wilcoxon1945,
  author    = "Wilcoxon, F.",
  title     = "Individual comparisons by ranking methods",
  journal   = "Biometrics Bulletin",
  volume    = "1",
  number    = "6",
  pages     = "80--83",
  year      = "1945",
  doi       = "10.2307/3001968"
}

@article{holm1979,
  author    = "Holm, S.",
  title     = "A simple sequentially rejective multiple test procedure",
  journal   = "Scandinavian Journal of Statistics",
  volume    = "6",
  number    = "2",
  pages     = "65--70",
  year      = "1979",
  note      = "\url{https://www.jstor.org/stable/4615733}"
}

\end{document}